%% file: final.tex
\documentclass[sigconf,11pt]{acmart}

\usepackage{booktabs} 
\usepackage{times}

\setcopyright{none}

\acmConference[6UAR'17]{MIT 6.UAR}{MAY 2017}{
  Cambridge, Massachusetts} 

\begin{document}
\title{False arrhythmia alarm reduction in the intensive care unit}

\author{Andrea S. Li}
\affiliation{%
  \institution{Massachusetts Institute of Technology}
}
\email{liandrea@mit.edu}

\author{Alistair E. W. Johnson}
\affiliation{%
  \institution{Massachusetts Institute of Technology}
}
\email{aewj@mit.edu}

\author{Roger G. Mark}
\affiliation{%
  \institution{Massachusetts Institute of Technology}
}
\email{rgmark@mit.edu}

\begin{abstract}
Research has shown that false alarms constitute more than 80\% of the alarms triggered in the intensive care unit (ICU). The high false arrhythmia alarm rate has severe implications such as disruption of patient care, caregiver alarm fatigue, and desensitization from clinical staff to real life-threatening alarms. A method to reduce the false alarm rate would therefore greatly benefit patients as well as nurses in their ability to provide care. We here develop and describe a robust false arrhythmia alarm reduction system for use in the ICU. Building off of work previously described in the literature, we make use of signal processing and machine learning techniques to identify true and false alarms for five arrhythmia types. This baseline algorithm alone is able to perform remarkably well, with a sensitivity of 0.908, a specificity of 0.838, and a PhysioNet/CinC challenge score of 0.756. We additionally explore dynamic time warping techniques on both the entire alarm signal as well as on a beat-by-beat basis in an effort to improve performance of ventricular tachycardia, which has in the literature been one of the hardest arrhythmias to classify. Such an algorithm with strong performance and efficiency could potentially be translated for use in the ICU to promote overall patient care and recovery. 
\end{abstract}

\maketitle

\input{body}

\bibliographystyle{abbrv}
\bibliography{bib}

\end{document}

%% file: body.tex
\section{Introduction}
False alarms are extremely common in the intensive care unit (ICU), and research has found that in certain settings, only 17\% of alarms are classified as clinically relevant \cite{siebig}. The resulting constant barrage of clinical alarms can lead to disruption of care and caregiver alarm fatigue, resulting in slower response time to real life-threatening events due to desensitization of clinical staff \cite{imhoff}. Alarm systems can also produce sound intensities of more than 80 decibels which can result in unwanted patient outcomes including sleep deprivation, inferior sleep structure, patient and provider stress, and depressed immune systems \cite{topf}. Thus, systems to reduce false alarm rates in the ICU have much potential to improve standard of care in the ICU and overall patient outcomes \cite{aboukhalil}. \\

A number of algorithms to try to reduce the false arrhythmia alarm rate have been investigated and developed to varying degrees of success in recent years \cite{plesinger}. In particular, as part of the 2015 PhysioNet/Computing in Cardiology (CinC) Challenge, participants were invited to specifically address the issue of high false alarm rates in the ICU \cite{plesinger}. \\ 

Plesinger \textit{et al.} \cite{plesinger} employed biomedical signal processing techniques to analyze the ECG, ABP, and PPG data channels for evidence of true and false alarms. By running a series of general and arrhythmia-specific tests, the algorithm was able to achieve an overall sensitivity of 0.93 and an overall specificity of 0.87 \cite{plesinger}. The algorithm described herein builds off of the algorithm developed by Plesinger \textit{et al.} , with a specific focus on optimizing for ventricular tachycardia, which exhibited the worst classification performance out of all the arrhythmias examined (sensitivity and specificity of 0.83) \cite{plesinger}. \\

Kalidas and Tamil \cite{kalidas} sought to solve the same problem but instead used support vector machine-based supervised learning for arrhythmia detection. Despite some acknowledged limitations in their approach, including the absence of PPG data and noise filtering, when scored against the PhysioNet/CinC Challenge scoring criterion, the algorithm ranked second after Plesinger with a challenge score of 0.794 \textit{et al.} \cite{kalidas,challenge}. \\

\begin{table*}
  \caption{Definitions of arrhythmias examined here}
  \begin{tabular}{cc}
    \toprule
    Arrhythmia                            & Formal definition \\
    \midrule
    Asystole                              & No QRS for 4 seconds \\
    Extreme bradycardia                   & Heart rate lower than 40 bpm for 5 consecutive beats \\
    Extreme tachycardia                   & Heart rate higher than 140 bpm for 17 consecutive beats \\
    Ventricular tachycardia               & Ventricular heart rate higher than 100 bpm for 5 consecutive ventricular beats \\
    Ventricular flutter/fibrillation      & Fibrillatory, flutter, or oscillatory waveform for 4 seconds \\
  \bottomrule
\end{tabular}
\end{table*}

Despite significant forward strides in false arrhythmia alarm detection explored by Plesinger \textit{et al.} \cite{plesinger}, Kalidas and Tamil \cite{kalidas}, etc., much work to decrease false positives and minimize false negative rates to 0\% must be done before false alarm algorithms can be safely used in a clinical setting. This paper therefore seeks to build off of the successful approaches of others' previous work, as well as to add on signal processing and machine learning techniques, to yield a more accurate and high fidelity algorithm and create improvements in patient care. \\

We aim to explore the use of advanced signal processing and machine learning techniques to reduce the rate of false arrhythmia alarms in the ICU while maintaining an extremely low false negative rate. Five arrhythmias in particular are examined here: asystole, extreme bradycardia, extreme tachycardia, ventricular fibrillation/flutter, and ventricular tachycardia, with the formal definitions for each outlined in Table 1 \cite{challenge}. Because false negatives are extremely costly and the false negative rate should arguably be exactly 0\% in a clinical setting, developing a clinically acceptable algorithm while improving upon the false positive rate presents unique challenges. \\

In the algorithm described herein, we make use of information from simultaneously measured electrocardiogram (ECG), arterial blood pressure (ABP), and photoplethysmogram (PPG) signals to reduce the impact of noise on a single channel. Signal quality metrics are used to inform how much data in each channel should be weighted in our modified voting algorithm, which is able to do a modest job at removing false alarms while introducing minimal false negatives. Such a false alarm reduction algorithm with high accuracy, sensitivity, and specificity can lead to a higher quality of care for patients in the ICU while maintaining rapid response time to real life-threatening situations. \\

\section{Methods} 
In order to reduce the false alarm rate for the five cardiac arrhythmias examined here, we reimplement an algorithm previously described in the literature and subsequently explore dynamic time warping to improve the approach and overall performance of the algorithm, specifically in relation to ventricular tachycardia. \\

\subsection{Dataset} 
We here use the dataset provided through the 2015 PhysioNet/CinC Challenge \cite{challenge}. This is a dataset of 750 true and false alarms of the five life-threatening cardiac arrhythmias examined here. These true and false alarm labels have been assigned by professional cardiologists after visual examination of the multichannel signal. The number of true and false alarms given for each arrhythmia are listed in Table 2. \\

\begin{table*}
  \caption{Rhythm true and false alarm counts}
  \label{tab:freq}
  \begin{tabular}{cccc}
    \toprule
    Arrhythmia    & \# total     & \# true alarms     & \# false alarms \\
    \midrule
    Asystole                              & 12           & 22                 & 100              \\
    Extreme bradycardia                   & 89           & 46                 & 43               \\
    Extreme tachycardia                   & 140          & 131                & 9                \\
    Ventricular tachycardia               & 341          & 89                 & 252              \\
    Ventricular flutter/fibrillation      & 58           & 6                  & 52               \\
  \bottomrule
\end{tabular}
\end{table*}

\subsection{Baseline algorithm}
We reimplement the algorithm described by Plesinger \textit{et al.} \cite{plesinger} in Python using a number of standard QRS detectors. An overview of this technical approach is given in Figure 1. Invalid segments in all data channels are first identified based on whether nominal values and signal variance are within expected limits (\textit{e.g.} blood pressure is positive but less than 300 mmHg), as well as through frequency analysis using band pass filters. Based on these calculations, the overall validity of a data channel is determined and the reliability of data in that channel is weighted accordingly. The results of standard QRS detectors, including GQRS and JQRS, yield QRS annotations which are used in denoting the locations of beats and in determining heart rate of the overall signal. \\

\begin{figure*}
\includegraphics[height=1.4in, width=6.5in]{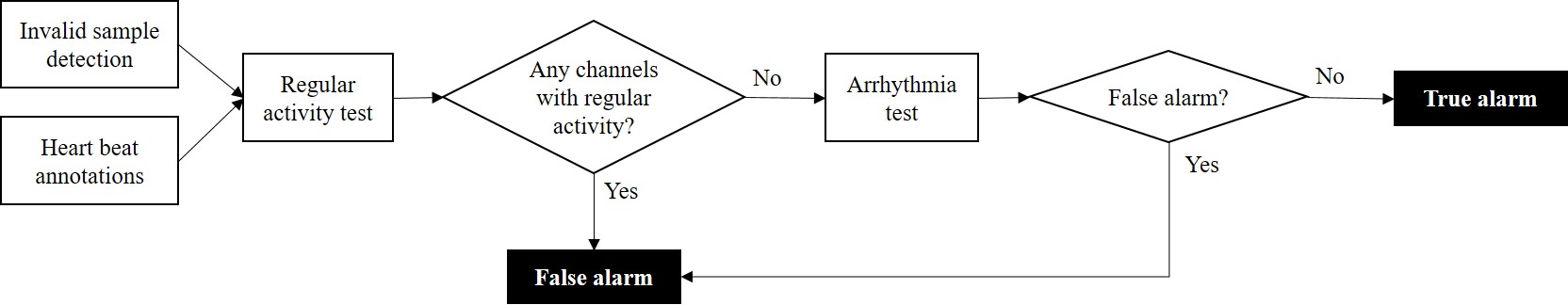}
\caption{Overview of baseline false arrhythmia alarm reduction algorithm.}
\end{figure*}

Information from invalid segment detection and heart beat annotations are combined in order to determine whether each channel exhibits regular activity in the vicinity of when the alarm was originally triggered. Regular activity is indicated if the signal does not contain invalid segments, as well as if the heart rate and RR intervals are regular, evenly distributed, and within reasonable limits. If any channel exhibits regular activity, the alarm is dismissed as noise and flagged off as a false alarm. Regular activity is here used as a way to invalidate alarms as false alarms because in real life-threatening situations, patients are physiologically unlikely to experience regular activity in any data channel. \\

If none of the channels indicate regular activity, arrhythmia-specific tests are run. The asystole test checks if there are any heart beats in a rolling three second window. The bradycardia test chooses the most reliable channel and compares the minimum heart rate for each window of four consecutive beats against a threshold of 45 beats per minute (bpm). The tachycardia test conversely compares the maximum heart rate for each window of 17 consecutive beats to a threshold of 140 bpm. The ventricular flutter/fibrillation test checks for fibrillatory or oscillatory behavior, characterized by a period of low frequency dominance. The ventricular tachycardia test analyzes both ECG channels and the ABP channel. For ECG channels, the test involves classifying each beat during the alarm signal as ventricular if the beat exhibits more low frequency energy than high frequency energy, or normal otherwise. The algorithm subsequently compares each window of four consecutive ventricular beats to a ventricular heart rate threshold of 95 bpm. For the ABP channel, the test compares the standard deviation of the ABP channel against a threshold of 6 mmHg, as during ventricular tachycardia the amplitude of the blood pressure generally falls significantly \cite{plesinger}. If the arrhythmia-specific test indicates a true alarm, the alarm is allowed, and otherwise the alarm is flagged off as a false alarm (Figure 1). \\

This initial approach closely mirrors that presented by Plesinger \textit{et al.} \cite{plesinger} but one major improvement on the algorithm has been made: in the algorithm as described by Plesinger \textit{et al.} \cite{plesinger}, in the ventricular tachycardia test, if multiple channels disagreed on whether a signal was a true alarm or not, the alarm was flagged off as a false alarm. In order to minimize the number of false negatives this approach introduces, we modify this to be such that if any channel detects ventricular tachycardia, then the alarm is allowed as a true alarm. \\

\subsection{Dynamic time warping}

Ventricular tachycardia has in the literature been one of the most challenging arrhythmias to classify \cite{aboukhalil}. As a result, we investigate dynamic time warping (DTW) as an alternative approach to identify ventricular tachycardia true and false alarms in particular. For all DTW approaches, the analysis was preliminarily only run on lead II of the ECG signal. If results are promising however, this approach can be extended to other leads of the ECG in a straightforward manner.\\

\subsubsection{Full alarm signal}
The dataset was first segmented into a training set of 500 multichannel signals and a testing set of 250 multichannel signals. These signals were all downsampled to 125 Hz from an original sampling frequency of 250 Hz. Furthermore, due to AAMI regulations \cite{standards}, an alarm must be raised within ten seconds of an arrhythmia event. As a result, only the last ten seconds before an alarm is triggered is analyzed as the caridac episode is guaranteed to occur sometime within this time frame. \\

Channels of a given test signal are normalized and dynamically time warped to every other channel of the same lead and rhythm type in the training set. This is performed with both a radius of 0 (regular Euclidean distance) and with radius of two seconds (250 samples for the downsampled signal at 125 Hz). Using $k$-nearest neighbors with $k=1$ on the Euclidean distance between warped channels, predictions are made on whether a given signal is a true alarm or false alarm. \\

\subsubsection{Beat-by-beat -- ventricular beat bank}

As the episode of ventricular tachycardia may occur anywhere in the alarm signal, we also investigate DTW on a beat-by-beat basis instead of on the entire ten-second signal. In this case, we classify each beat as a ventricular beat or as a non-ventricular beat. This classification is subsequently used in the baseline algorithm to try to boost performance of classification of ventricular tachycardia alarms. \\

We manually compile a set of 20 ventricular beats from patients who experienced true ventricular tachycardia episodes and store these in a "representative" beat bank. In addition, we extract 20 non-ventricular beats from the patient before the alarm section and store these in a "standard" beat bank (Figure 2). \footnote{Here, a beat is na\"ively denoted as a third of the way before the beat annotation to the prior annotation until two-thirds after the beat annotation to the next annotation.} These non-ventricular beats are only included in the bank of representative non-ventricular beats if the ten-second section of the signal in which the beat is located exhibited high quality. Requirements for high quality and clean signals are given in Table 3. \\

\begin{figure}
\includegraphics[height=2.2in, width=2.5in]{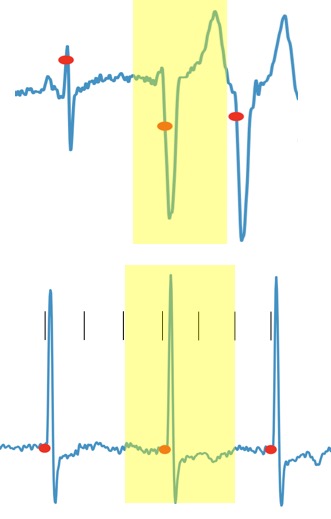}
\caption{Ventricular beats manually identified (top) vs. non-ventricular patient-specific beats automatically extracted (bottom)}
\end{figure}

\begin{table}
  \caption{Requirements for clean signals for extraction of self non-ventricular beats}
  \label{tab:freq}
  \begin{tabular}{ccc}
    \toprule
    Metric    & Definition     & Threshold value \\
    \midrule
    Baseline wander  & $1-\frac{\int_{f=0}^{f=1}PSD df}{\int_{f=0}^{f=40}PSD df}$   & 0.75  \\
    Power            & $\frac{\int_{f=5}^{f=15}PSD df}{\int_{f=5}^{f=40}PSD df}$    & 0.9   \\
    Kurtosis         & $\frac{1}{M}\sum_{i=1}^M [\frac{x_i - \mu_x}{\sigma}]^4$     & 4     \\
  \bottomrule
\end{tabular}
\end{table}

To classify beats during the alarm segment as ventricular or non-ventricular, we run DTW with a one-second radius for each detected beat against the beat bank. $k$-nearest neighbors with $k=1$ is again used to choose the signal with the smallest warped Euclidean distance in order to determine if a detected beat is a non-ventricular beat or a ventricular beat. These beat annotations are used in the baseline algorithm to substitute for the current ventricular beat detector. \\ 

\subsubsection{Beat-by-beat -- self-beat bank}

As ventricular beats can vary significantly from patient to patient, a universal set of standard or representative ventricular beats may not necessarily exist. We therefore additionally investigate an alternative approach in which we only compare patient beats with previous beats from the same patient. This approach is based on the assumption that the patient's rhythm prior to the alarm is normal (\textit{i.e.} not ventricular tachycardia). \\

As before, we identify a set of 20 standard beats from before the alarm signal using sections of high quality signal determined by the metrics presented in Table 3. In contrast to the ventricular beat bank approach of before, we approach ventricular beat classification as a novelty detection problem.
We establish the "expected" distance of a new non-ventricular beat by calculating the pairwise distances of each beat within the beat bank. We then calculate the minimum distance for each beat, resulting in 20 minimum distance values. We define $\mu_{i, min}$ as the mean of these values and $\sigma_{i, min}$ as the standard deviation. \\

In the alarm signal, each unknown beat is warped to each of the non-ventricular beats in the bank of 20 self-beats. If the minimum warped distance is greater than $\mu_{i, min} + \sigma_{i, min}$, then the beat is denoted as a ventricular beat. Otherwise, the beat is denoted as a non-ventricular beat (Figure 4). \\

\begin{figure}
\includegraphics[height=1.1in, width=3in]{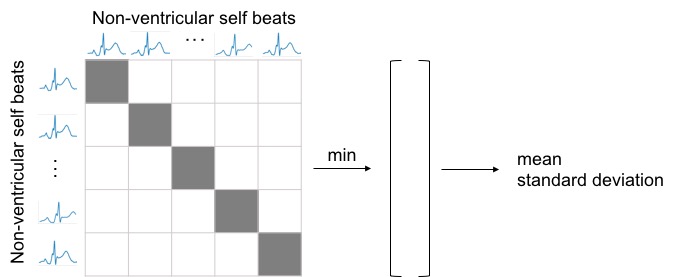}
\caption{Non-ventricular self beats}
\end{figure}


\begin{figure}
\includegraphics[height=1.1in, width=3in]{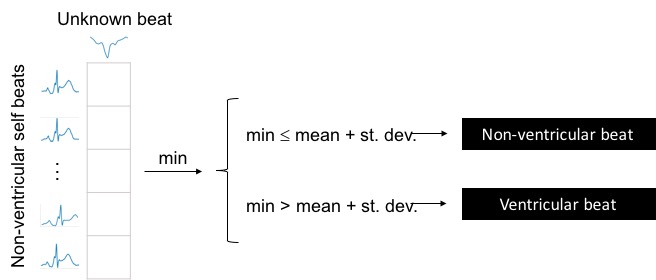}
\caption{Determining whether new beats are ventricular or not.}
\end{figure}

In addition to exploring the minimum warped distance found for each test signal, we also use the Kullback-Leibler (KL) divergence as a measure of difference between the distributions of warped distances. This measure of similarity is given by: 

\begin{equation} 
  D_{KL}(P|Q) = \sum_i P(i) \log \frac{P(i)}{Q(i)}
\end{equation}

The mean $\mu_{i, KL}$ and standard deviation $\sigma_{i, KL}$ for KL divergence are found as with the minimum distance metric. Thus, if the KL divergence value for a test signal is greater than $\mu_{i, KL} + \sigma_{i, KL}$, then the beat is classified as a ventricular beat. Otherwise, the beat is classified as a non-ventricular beat. These beat annotations are again utilized in the baseline algorithm to substitute for the current ventricular beat detector. \\

\subsection{Evaluation metrics}
In order to evaluate the performance of the baseline algorithm and DTW approaches, we calculate and compare several quantitative metrics with those of existing algorithms. Evaluation metrics include sensitivity, specificity, positive predictive value (PPV), negative predictive value (NPV), F1 (the harmonic mean of PPV and sensitivity), and the scoring metric from the 2015 PhysioNet/CinC Challenge calculated as: 
\begin{equation}
  Score = \frac{TP+TN}{TP+TN+FP+5*FN}
\end{equation}
where $TP$, $TN$, $FP$, and $FN$ are the number of true positives, true negatives, false positives, and false negatives \cite{challenge}. Evaluation metrics are calculated overall for all five arrhythmia types, as well as individually for each arrhythmia class. \\

\section{Results and discussion}
We seek to minimize the false arrhythmia alarm rate in the ICU while simultaneously introducing a near-0 number of false negatives. We reimplement an existing algorithm to tackle this problem and seek to improve performance for ventricular tachycardia in particular through dynamic time warping techniques. We anticipate the performance of the baseline algorithm reimplementation to be comparable with the results of Plesinger \textit{et al.} \cite{plesinger}, with potentially improved performance metrics resulting from the time warping techniques. \\

\subsection{Baseline algorithm}
Overall, the baseline algorithm with the improvement outlined above was able to achieve a sensitivity of 0.908 and a specificity of 0.838 for all arrhythmia types examined here. This yielded a Physionet/CinC challenge score of 0.756. Furthermore, the improved baseline algorithm was able to reduce the false alarm rate to 9.9\%, while introducing 3.6\% false negatives. \\

This algorithm was additionally evaluated for each arrhythmia type with different QRS detectors published in the literature. In Table 4, these results are compared with the results of the improved baseline algorithm and the results presented by Plesinger \textit{et al.} \cite{plesinger}, where n/a indicates that the value was not reported. Overall, the results of the improved baseline algorithm are quite comparable to the results of the algorithm presented by Plesinger \textit{et al.} \cite{plesinger}. \\

\begin{table*}
  \caption{Performance of each QRS detector, improved baseline algorithm, and Plesinger \textit{et al.} algorithm}
  \begin{tabular}{cccccc}
    \toprule
      & & GQRS & JQRS & Improved algorithm & Plesinger \textit{et al.} \\
    \midrule
                            & Sensitivity       & 1.0   & 1.0   & 1.0   & 0.96 \\
                            & Specificity       & 0.92  & 0.83  & 0.93  & 0.92 \\
    \textbf{Asystole}       & Challenge score   & 0.934 & 0.861 & 0.943 & n/a  \\
                            & NPV               & 0.733 & 0.564 & 0.759 & n/a  \\
                            & PPV               & 1.0   & 1.0   & 1.0   & n/a  \\
                            & F1                & 0.826 & 0.721 & 0.863 & n/a  \\
    \midrule      
                            & Sensitivity       & 0.826 & 0.804 & 0.826 & 0.97 \\
                            & Specificity       & 0.953 & 0.953 & 0.953 & 0.74 \\
    \textbf{Bradycardia}    & Challenge score   & 0.653 & 0.624 & 0.653 & n/a  \\
                            & NPV               & 0.95  & 0.949 & 0.95  & n/a  \\
                            & PPV               & 0.837 & 0.82  & 0.837 & n/a  \\
                            & F1                & 0.653 & 0.624 & 0.653 & n/a  \\
    \midrule
                            & Sensitivity       & 0.985 & 1.0   & 0.985 & 1.0  \\
                            & Specificity       & 0.778 & 0.889 & 0.778 & 0.89 \\
    \textbf{Tachycardia}    & Challenge score   & 0.919 & 0.993 & 0.919 & n/a  \\
                            & NPV               & 0.985 & 0.992 & 0.985 & n/a  \\
                            & PPV               & 0.778 & 1.0   & 0.778 & n/a  \\
                            & F1                & 0.919 & 0.993 & 0.919 & n/a  \\
    \midrule      
                            & Sensitivity       & 0.921 & 0.831 & 0.820 & 0.83 \\
                            & Specificity       & 0.480 & 0.639 & 0.786 & 0.83 \\
    \textbf{Ventricular}    & Challenge score   & 0.550 & 0.586 & 0.669 & n/a  \\
    \textbf{tachycardia}    & NPV               & 0.385 & 0.448 & 0.575 & n/a  \\
                            & PPV               & 0.945 & 0.915 & 0.925 & n/a  \\
                            & F1                & 0.543 & 0.583 & 0.676 & n/a  \\
    \midrule
                            & Sensitivity       & 1.0   & 1.0   & 1.0   & 1.0  \\
                            & Specificity       & 0.904 & 0.904 & 0.904 & 0.90 \\
    \textbf{Ventricular}    & Challenge score   & 0.914 & 0.914 & 0.914 & n/a  \\
    \textbf{flutter/fibrillation} & NPV         & 0.545 & 0.545 & 0.545 & n/a  \\
                            & PPV               & 1.0   & 1.0   & 1.0   & n/a  \\
                            & F1                & 0.914 & 0.914 & 0.914 & n/a  \\
  \bottomrule
\end{tabular}
\end{table*}

Performance of the baseline algorithm in classifying bradycardia and ventricular tachycardia were generally much worse than that of classifying asystole, tachycardia, and ventricular flutter/fibrillation. This is especially seen in the breakdown of false negatives generated by the improved baseline algorithm as illustrated in Figure 5. We therefore seek to focus more specifically on the classification task of identifying ventricular tachycardia alarms as true or false alarms\footnote{We chose not to focus on bradycardia as the reduced performance was primarily due to errors in the QRS detection, which was not the focus of our work.}. \\

\begin{figure}
\includegraphics[height=1.7in, width=3in]{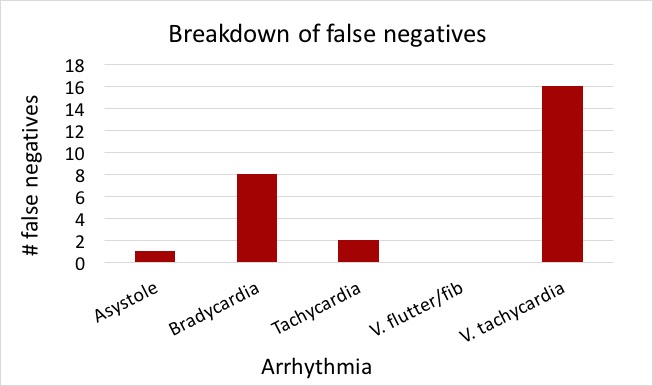}
\caption{Breakdown of false negatives in the dataset of 750 arrhythmias of all five arrhythmia types.}
\end{figure}

\subsection{Dynamic time warping}

Because all dynamic time warping approaches were preliminarily investigated for a single lead for ventricular tachycardia, all results presented in the following are based on single-lead (usually lead II) results. Comparisons with the baseline algorithm are therefore given against the results of the baseline algorithm assuming a single ECG signal instead of a multimodal approach. \\

Interestingly, when isolated to a single lead, the improved baseline algorithm performed worse than the original baseline algorithm. Thus, it is worth noting that robust performance in a single lead does not necessarily directly translate to robust performance overall in a multimodal setting. \\

\subsubsection{Full alarm signal}

DTW on the full alarm signal generated slightly worse results compared with the baseline algorithm for a radius of 0. More specifically, DTW with a radius of 0 (Euclidean distance) resulted in a sensitivity of 0.515, a specificity of 0.774, and a PhysioNet/CinC challenge score of 0.453. This is compared with a challenge score of 0.508 for the baseline algorithm and a score of 0.404 for the improved baseline algorithm. \\

However, DTW on the full alarm signal with a radius of two seconds resulted in better performance compared with both the baseline algorithm and the improved baseline algorithm. In this case, the DTW algorithm resulted in a sensitivity of 0.567, a specificity of 0.798, and a challenge score of 0.515. This performance is only slightly better than the baseline score of 0.508 but significantly better than the improved baseline score of 0.404. \\

The lower sensitivity and higher specificity metrics seen here indicate that the skewed loss function in the clinical setting, in which false negatives are penalized higher than false positives, is not inherently optimized for by the algorithm. In contrast, the improved baseline algorithm generally skews towards an improved sensitivity and lower specificity to account for the increased penalty for a false negatives. \\ 

Furthermore, the run of ventricular beats is not necessarily in the same location in each alarm signal. As a result, even two true ventricular tachycardia episodes might result in high warped distances if the episodes are located at different locations in the signal. This may have contributed to the fact that DTW on the alarm signal only generated slightly improved performance results compared with the baseline algorithm. \\

\subsubsection{Beat-by-beat -- ventricular beat bank}

Dynamic time warping on a beat-by-beat basis with a ventricular beat bank performed slightly better than the baseline algorithm and DTW on the entire alarm signal. The single-channel sensitivity was 0.767, the single-channel specificity was 0.598, and the challenge score was 0.518. \\

This approach to identification of ventricular beats may not have resulted in as significant of an improvement as expected because there is not necessarily a representative morphology for a ventricular beat. Thus, ventricular beats can be true beats but still not be recognized as ventricular beats if the morphology of the beat had not previously been seen in other patients. This approach is therefore significantly dependent on the ventricular beats available in the ventricular beat bank, which may not necessarily capture all morphologies of ventricular beats. Thus, while this approach resulted in slight performance improvements over the baseline algorithm, significant improvements were not seen potentially due to the patient-specific nature and morphology of ventricular beats. \\

\subsubsection{Beat-by-beat -- self-beat bank}

We also investigated the use of beat-by-beat DTW against a bank of extracted non-ventricular beats from the patient in question. With the KL divergence as a measure of the similarities and differences between beats, this generated worse performance compared with both the baseline algorithm and the improved baseline algorithm. More specifically, a sensitivity of 0.116, a specificity of 0.960, and a challenge score of 0.390 resulted from this approach. Thus, although specificity was extremely high, the low sensitivity number indicates a high number of false negatives, which are punished heavily in the challenge score. \\

This approach may have performed so poorly due to the importance that potentially incorrectly detected beats held in the algorithm. More specifically, if QRS annotations were missing or incorrectly detected by the QRS detector, this could result in dramatically shorter or longer beats included in the bank of self non-ventricular beats. As the KL divergence measure takes all the distances into account, these incorrect beats could have skewed the distribution of distances for a new test signal, as well as incorrectly impacted the mean and standard deviation of the pairwise distance comparisons. \\

In contrast, using the minimum distance as the metric for similarities and differences to non-ventricular beats resulted in improved performance compared with KL divergence, but still worse performance compared with the baseline algorithm. This approach likely saw better performance compared with KL divergence, as the impact of incorrect beats is reduced by only looking at the minimum distance, instead of a measure which incorporates all distances. \\


Poor performance of this ventricular beat detector compared with the baseline algorithm could be due to the fact that in this approach, beats are only compared with non-ventricular beats. Thus, both noise and ventricular beats in the alarm signal are treated similarly (both result in large distances from non-ventricular beats), but there exists no mechanism to distinguish between ventricular beats and noise. \\

The results for all DTW approaches compared with the baseline algorithm and improved baseline algorithm are presented in Table 5. These results represent analysis of only ventricular tachycardia with a single lead (usually lead II) for consistency in comparison of performances. \\

\begin{table*}
  \caption{Performances for ventricular tachycardia (single lead)}
  \begin{tabular}{ccccccc}
    \toprule
      & Sensitivity & Specificity & PPV & NPV & F1 & Challenge score \\
    \midrule
      Baseline algorithm                            & 0.744  & 0.606 & 0.395 & 0.873 & 0.516 & 0.508 \\
      Improved baseline algorithm                   & 0.535  & 0.618 & 0.326 & 0.794 & 0.405 & 0.404 \\
      DTW on alarm signal, radius=0                 & 0.515  & 0.774 & 0.472 & 0.802 & 0.493 & 0.453 \\
      DTW on alarm signal, radius=250 (2 seconds)   & 0.567  & 0.798 & 0.486 & 0.845 & 0.515 & 0.515 \\
      DTW beat-by-beat, ventricular beat bank       & 0.767  & 0.598 & 0.398 & 0.882 & 0.524 & 0.518 \\
      DTW beat-by-beat, self-beat bank (KL)         & 0.116  & 0.960 & 0.500 & 0.759 & 0.189 & 0.390 \\
      DTW beat-by-beat, self-beat bank (min)        & 0.535  & 0.655 & 0.349 & 0.803 & 0.422 & 0.422 \\
    \bottomrule
\end{tabular}
\end{table*}

\subsection{Feasibility for real-time algorithm}

In terms of runtime, the baseline algorithm and beat-by-beat DTW are likely most feasible for a real-time setting, in which decisions on true and false alarms must be made rapidly. However, in terms of adaptability and extensibility, the beat-by-beat DTW with a ventricular beat bank suffers somewhat. This is because the ventricular beats in the ventricular beat bank must be manually identified and added to the bank. Furthermore, at a larger scale, the ventricular beat bank approach might suffer significantly, as the number of different types and morphologies of ventricular beats increases, while the number of beats in the ventricular beat bank stays constant. Semi-supervised parametric methods for identifying ventricular beats may alleviate this issue to some degree \cite{oster2015semisupervised}. \\

Thus, the improved baseline algorithm as proposed would likely fare best in a real-time setting. Alternatively, the DTW beat-by-beat self-beat approach has significant room for improvement. If this approach is optimized to perform better than the baseline algorithm, then the self-beat approach would likely be more scalable than the ventricular beat bank approach, as the self-beat approach is not dependent on a small set of manually extracted ventricular beats. \\

\section{Conclusions and future work}
We here describe and implement a system to reduce the false arrhythmia alarm rate in the intensive care unit by analyzing multimodal data using signal processing and machine learning techniques. By building off of previous work, the improved baseline algorithm run on multimodal data was able to achieve an overall sensitivity of 0.908, specificity of 0.838, and a PhysioNet/CinC challenge score of 0.756. We additionally investigate the use of dynamic time warping and $k$-nearest neighbors with $k=1$ on the ten seconds prior to an alarm to classify overall signals, as well as on a beat-by-beat basis to identify ventricular beats and non-ventricular beats. These algorithms have the potential to generate slightly improved performances over the baseline algorithm, with the beat-by-beat DTW algorithm using a ventricular beat bank resulting in the best performance. These optimized algorithms in the future, with minimal false negatives, could significantly and positively impact patient care, provider stress, and overall expected health outcomes in the hospital. \\

The dynamic time warping approaches especially have significant room for improvement in the future. In particular, incorporating multiple channels in DTW analysis will likely provide a significant source of improvement. Beyond this, beat-by-beat DTW with patient self-beats can likely experience performance benefits through improved beat identification and optimization of parameters, such as thresholds for clean versus noisy signals, cutoff thresholds for ventricular versus non-ventricular beats (currently at one standard deviation above mean), etc. \\

Furthermore, the self non-ventricular beats can currently be extremely skewed if annotations are incorrect (extra or missing annotations). This can pollute the warped distances, and by extension, the annotations of non-ventricular and ventricular beats. Thus, optimizations which seek to improve beat detection and filter out incorrect beats can potentially yield performance improvements. Overall, much future work exists to improve the algorithms described herein to maximally minimize the number of false negatives for ventricular tachycardia and other arrhythmias. \\

Prior to investigating whether such a system can be deployed in hospitals, the performance and accuracy of this false alarm algorithm must be optimized to clinically acceptable levels. In particular, the algorithm must be able to run fast enough to keep up in a real-time setting. Furthermore, accuracy of true and false alarm classification should be improved such that the number of false negatives approaches 0. Following these improvements, extensive testing of the algorithm must be done for a large number of patients at different levels of severity in many different ICUs with different equipment, systems, and standards. Pending successful results of these tests, the algorithm may be rolled out to hospitals and ICUs to reduce the false alarm rate and improve patient care in a clinical setting. \\
